\documentclass[conference]{IEEEtran}
\usepackage{cite}
\usepackage{amsmath,amssymb,amsfonts}
\usepackage{algorithmic}
\usepackage{graphicx}
\usepackage{textcomp}

\usepackage[english]{babel}
\usepackage[T1]{fontenc}
\usepackage{epsfig}
\IEEEoverridecommandlockouts
\usepackage{tabu}
\usepackage{float}
\usepackage{amsmath}
\usepackage{amssymb}
\usepackage{amsthm}
\usepackage[keeplastbox]{flushend}
\usepackage{cite}
\usepackage{subfigure}
\usepackage{epstopdf}
\usepackage{multirow}
\usepackage{caption}
\usepackage[table]{xcolor}
\usepackage{mathtools}
\usepackage{xfrac}
\usepackage{units}
\usepackage[hyphens]{url}
\usepackage{array}
\newcommand{\beql}[1]{\begin{equation}\label{#1}}
\newcommand{\eeq}{\end{equation}}

\newcommand{\be}{\begin{equation}}
\newcommand{\ee}{\end{equation}}
\newcommand{\ba}{\begin{array}}
\newcommand{\ea}{\end{array}}

\ifCLASSINFOpdf
\else
\fi
\begin{document}
%

\title{Impact of Learning Rate on Noise Resistant Property of  Deep Learning Models}
%
%
%
\author{Omobayode~Fagbohungbe,~\IEEEmembership{Student~Member,~IEEE,}
        ~Lijun~Qian,~\IEEEmembership{Senior~Member,~IEEE}
\thanks{The authors are with the CREDIT Center and the Department
of Electrical and Computer Engineering, Prairie View A\&M University, Texas A\&M University System, Prairie View, 
TX 77446, USA. Corresponding author: Omobayode Fagbohungbe e-mail: ofagbohungbe@pvamu.edu}}

\maketitle


\begin{abstract}
The interest in analog computation has grown tremendously in recent years due to its fast computation speed and excellent energy efficiency, which is very important for edge and IoT devices in the sub-watt power envelope for deep learning inferencing. However, significant performance degradation suffered by deep learning models due to the inherent noise present in the analog computation can limit their use in mission-critical applications. Hence, there is a need to understand the impact of critical model hyperparameters' choice on the resulting model noise-resistant property. This need is critical as the insight obtained can be used to design deep learning models that are robust to analog noise. In this paper, the impact of the learning rate, a critical design choice, on the noise-resistant property is investigated. The study is achieved by first training deep learning models using different learning rates. Thereafter, the models are injected with analog noise and the noise-resistant property of the resulting models is examined by measuring the performance degradation due to the analog noise. The results showed there exists a sweet spot of learning rate values that achieves a good balance between model prediction performance and model noise-resistant property. Furthermore, the theoretical justification of the observed phenomenon is provided.
\end{abstract}


\IEEEpeerreviewmaketitle

\section{Introduction}
\label{sec:Introduction}

Deep neural network (DNN) models continue to enjoy wide application in very challenging and complex real-time cognitive applications such as machine translation, autonomous driving, object detection, recommender systems, etc., due to their superlative ability to detect and recognize complex features in text, images, and speech. The inherently scalable nature of this ability of DNN models has enabled the design and deployment of very large models in commercial and industrial application~\cite{Burrgeo2021,Onasami2022UnderwaterAC,Yang2022AJE,Fagbohungbe2021EfficientPP}. Despite their impressive success, DNN models require millions of multiply and accumulate operations that consumes a lot of computing and memory resources. This requirement makes them unattractive for deployment in application areas with power envelopes of sub-Watt or of very few Watts, which are generally referred to as the IoT or edge computing realm~\cite{Kariyappa,dazzi2021, NoisyNN}. Although several methods such as low precision quantization techniques and weight prunning~\cite{han2015deep,luo2017thinet} methods have been introduced to reduce the arithmetic complexity and memory requirement of DNN models and many digital accelerators proposed, the progress has been very limited due to the von-Neumann nature of the platforms these models are deployed on~\cite{Kariyappa,dazzi2021, NoisyNN}. The von-Neumann nature means there is a continuous need for data movement and communication between the memory and compute unit, leading to high latency and high energy cost~\cite{Kariyappa,dazzi2021}. Hence, there is a need for hardware accelerators that are fast, reliable, and energy-efficient as an alternative to these von Neumann digital hardware~\cite{NoisyNN,joshi}.

This requirement has led to significant interest in analog accelerators, a form of analog computation which uses in-memory computing~\cite{Kariyappa,NoisyNN}. The interest in these accelerators can be attributed to the fact that they can potentially help perform matrix-vector multiplication, a significant computation load in DNN, with unprecedented speed and energy efficiency~\cite{dazzi2021, Fasoli2020, Kariyappa}. Leveraging on the physical characteristics of these resistant-based or charge-based memory devices and arranging them in a crossbar array configuration, a matrix-vector multiplication can be carried out with $O(1)$ time complexity using Ohm's and Kirchhoff's laws, compared to the $O(N^2)$ time complexity in digital accelerators~\cite{dazzi2021, Kariyappa, Fasoli2020}. With the computation happening in the memory unit, data transfer is substantially reduced, data parallelism is increased, and significant improvement in energy efficiency is achieved~\cite{Kariyappa,dazzi2021,Chen2020}.

However, computation in analog accelerators is very imprecise due to the inherent presence of noise or variability such as device-to-device variability, write noise, and conductance drift~\cite{dazzi2021,Chen2020,Kariyappa}. This noise can severely impact the performance of DNN models, as demonstrated in~\cite{Fagbohungbe2020BenchmarkingIP,NoisyNN}. Therefore, there is a need to design DNN models with excellent noise-resistant property to minimize the impact of these noise on the model performance. To achieve this, there is a need to investigate and understand the impact of various model hyperparameters on the noise-resistant property of DNN models. This is critical as this could help maintain the delicate balance between the model inference performance and its noise-resistant property. Although many methods have been proposed to improve the noise-resistant property of DNN models~\cite{joshi,NoisyNN,Kariyappa}, the majority of these methods did not investigate the impact or leverage on model hyperparameters to achieve these goals. Hence, this research seeks to know if model hyperparameters affect a model noise-resistant property, and if they can be explored to design models that are more resistant to noise. Specifically, the impact of the learning rate on the resulting model noise-resistant property is investigated in this paper. The choice of learning rate is informed by its significant importance in the model training process as it determines if the model training process is going to converge and also the speed of convergence.

Noise, either those inherently present in the training process or externally injected during the training process, has played a critical role in designing DNN models by helping the training algorithm converge to or close to the global minima. In fact, one of the popular methods to improve the noise-resistant property of DNN model is the noise injection method, which is injecting either the model input, model weight, or gradient of the model weight with noise during the forward propagation during training alone or in combination with other methods like knowledge distillation~\cite{joshi,NoisyNN,Kariyappa}. In fact, the default noise-resistant property of DNN models can be attributed to the inherent presence of noise in the training process in the form of an error term that has zero mean~\cite{Bjorck2018UnderstandingBN,Merolla2016DeepNN}. The zero mean value of the noise is due to the constant batch size during the training process. The error term is mathematically defined below in equation~(\ref{eequ4}) where $\alpha$ is the learning rate, $\nabla$ is the gradient, $ L_D(x)$ is the loss function over all the dataset, $ L_i(x)$ is the loss function for a single data point $i$, $\nabla L_{SGD}(x)$ is the estimated true loss gradient, and $B$ is the Batch size. This error term introduces some noise into the computed gradient of the weight, hence influencing the weights and properties of the resulting model. The upper bound of the noise of the gradient due to the error term is defined in equation (\ref{eequ7}) where $C$ is given in equation (\ref{eequ6})~\cite{Bjorck2018UnderstandingBN}.

\begin{equation}
\label{eequ4}
\alpha \nabla_{SGD}(x)= \alpha \nabla L_D(x) + \frac{\alpha}{B}{\sum_{i \in B} ({\nabla L_i(x) - \nabla L_D(x)})}
\end{equation}

\begin{equation}
\label{eequ7}
E[\|{{\alpha\nabla L_B(x) - \alpha\nabla L_{SGD}(x)}}\|^2] < \frac{\alpha^2}{|B|}{C}
\end{equation}

\begin{equation}
\label{eequ6}
C=E[\|{{\nabla L_i(x) - \nabla L_D(x)}}\|^2]
\end{equation}

It can be inferred that the learning rate affects the power of the inherent noise present in the training process, which in turn influences the noise-resistant property of the resulting DNN model. Hence, there is a range of learning rate values that strikes the delicate balance between model inference performance and the model noise-resistant property. With other factors fixed, the learning rate value below the sweet spot reduces the power of the inherent noise, which potentially can lead to a model with low noise resistant value, long training time, and poor inference performance. Learning rate values greater than the sweet spot increase the power of the inherent noise, leading to a model with poor noise-resistant property and excellent model inference performance. Therefore, selecting the appropriate learning rate value influences the gradient noise, resulting in DNN models of excellent noise-resistant property and outstanding inference performance.

To validate the postulations above, DNN models are trained with different learning rates, and the noise-resistant property of the resulting model is measured. The noise-resistant property is measured by injecting additive noise into all the model layers, and new inference accuracy due to the noise is measured and normalized using the model inference accuracy in the absence of noise. The normalized accuracy represents the degradation due to the injected noise. This paper makes the following contributions:

\begin{enumerate}
\item Provided theoretical analysis and insights on how learning rate affects the noise-resistant property of DNN models;
\item Established that there is a range of learning rate values that maintains the delicate balance between model inference performance and model noise-resistant property;
\item Discovered that for a fixed learning rate, the choice of model optimizer would impact the noise-resistant property of the resulting DNN model.
\end{enumerate}

The remainder of this paper is organized as follows: The detailed proposed methodology for this work is discussed in Section~\ref{sec:methods}. Results and analysis are presented in Section~\ref{sec:results}. Further discussions and related works are reviewed in Section~\ref{sec:discussion} and Section~\ref{sec:conclusion} concludes the paper.

\begin{table*}
\centering
\caption{The details of the DNN models and dataset used in the experiments.}
\label{table:details}
\begin{tabular}{|*{6}{c|} }
\hline
Model Name & Dataset & Number of Classes & Model Input Dimension & \# Images per class\\
\hline\hline
ResNet20 & [CIFAR10, CIFAR100] & [10,100] & 32*32*3 & [6000,600] \\
\hline
ResNet32 & [CIFAR10, CIFAR100] & [10,100] & 32*32*3 & [6000,600]\\
\hline
ResNet44 & [CIFAR10, CIFAR100] & [10,100] & 32*32*3 & [6000,600] \\
\hline
ResNet56 & [CIFAR10, CIFAR100] & [10,100] & 32*32*3 & [6000,600]\\
\hline
\end{tabular}
\end{table*}

\section{Proposed Methodology}
\label{sec:methods}

This section details the procedures followed in performing the various experiments to study the impact of the learning rate on the noise-resistant property of the DNN model. The discussion covers various headings such as dataset used, model design and training, software and hardware used.

\subsection{Dataset}
\label{subsec:ExpSetup}

The CIFAR10 and CIFAR100 dataset~\cite{krizhevsky2009learning}, designed, compiled, and labeled by the Canadian Institute for Advanced Research out of the 80 million tiny images dataset, is used in this work. The datasets are very similar in that they contain colored images of dimension $32\times32\times3$ and are used for image classification tasks. Furthermore, they both contain 60,000 colored images, which can be sub-divided into 50,000 training images and 10,000 testing/inference images. The difference lies in that the CIFAR10 dataset contains images that can be grouped into ten mutually exclusive classes without any semantic overlap between the classes. However, the CIFAR100 contains images that can be grouped into 100 non-mutually exclusive classes with some form of semantic overlap as the dataset contain just 20 superclasses.

\subsection{Model Design and Training}
\label{subsec:ExpSetup}
ResNet~\cite{resnet} model of various depths trained on CIFAR10 and CIFAR100 datasets as detailed in Table~\ref{table:details} is used in this work. ResNet models are a form of convolutional neural network based on the residual learning framework, which eases the training and optimization of deeper models. It achieves this by reformulating the constituent layers of a model as a learning residual function with reference to the layers inputs instead of learning unreferenced function~\cite{resnet}. The choice of ResNet models is influenced by their popularity and excellent performance on various classification tasks of varying complexity. Furthermore, the models used in this work are trained from scratch until convergence is achieved by minimizing the cost function, which is the categorical cross-entropy. The cost function minimizes the error between the ground truth and the predicted label. This is achieved using the glorot-uniform method as an initializer, Adams optimization algorithm as the optimizer, and performing Data augmentation during training to prevent overfitting.

\begin{table*}
\centering
\caption{Comparison of the baseline inference accuracy of DL models trained with learning rate of 0.0005, 0.000625, 0.00075, 0.001, 0.00125 and 0.00150 using CIFAR10 and CIFAR100 dataset without noise injection to the weights of the trained model during testing. The performance metric is the model classification accuracy.}
\label{tab:result_baseline}
\begin{tabular}{|c|c|c|c|c|c|c|c|c|c|c|c|c|c|c|c|c|c|c|c|c|c|}
\hline



{Dataset}& \multicolumn{6}{|c|}{CIFAR10}& \multicolumn{6}{|c|}{CIFAR100}\\ \hline

& \multicolumn{6}{|c|}{Learning Rate}& \multicolumn{6}{|c|}{Learning Rate}\\ \hline
{Model Name} & {0.0005} & {0.000625} & {0.000725}& {0.001} & {0.00125} & {0.0015} &
{0.0005} & {0.000625} & {0.000725}& {0.001} & {0.00125} & {0.0015}\\ \hline

{Resnet\_20} & {91.11\%} & {91.39\%}& {91.87\%}& {91.27\%} & {91.53\%}& {91.49\%}
& {65.22\%} & {66.05\%} & {66.79\%}& {66.89\%} & {67.37\%} & {68.37\%}\\ \hline

{Resnet\_32} & {91.51\%} & {92.13\%}& {91.18\%}& {91.72\%} & {89.471\%}& {91.54\%}
& {67.78\%} & {68.27\%} & {68.60\%}& {69.39\%} & {68.70\%} & {67.68\%}\\ \hline

{Resnet\_44} & {92.11\%} & {92.63\%}& {92.26\%}& {92.24\%} & {91.62\%}& {91.23\%}
& {68.69\%} & {69.36\%} & {70.07\%}& {69.38\%} & {69.57\%} & {68.77\%}\\ \hline

{Resnet\_56} & {92.24\%} & {92.62\%}& {92.39\%}& {91.94\%} & {91.75\%}& {91.72\%}
& {69.10\%} & {69.93\%} & {70.01\%}& {69.79\%} & {69.82\%} & {68.79\%}\\ \hline

\hline
\end{tabular}
\end{table*}

\subsection{Model Training Stage}
\label{sec:training}
The four DL models, namely, ResNet20, ResNet34, ResNet44 and ResNet56 are trained from scratch using six different learning rates (0.0005, 0.000625, 0.00075, 0.001, 0.00125 and 0.00150) until convergence. This results in 24 trained models for each dataset.

\subsection{Model Inference stage}
\label{sec:inference}
The performance of the various models trained in section \ref{sec:training} in the absence and presence of noise is evaluated in this section using the appropriate metrics in order to understand the impact of learning rate on model noise-resistant property. The performance metric of the models in this work is the model inference accuracy (\%), as all the models under consideration are classification models. The inference accuracy of the models in the absence of noise is the baseline inference value of the models, and it is essential in order to calculate the performance degradation due to the presence of noise measured using the normalized inference accuracy defined in equation (\ref{equ4}).

To measure the performance of the models in the presence of noise, noise is injected into all the model layers. The noise in this work is modeled as white additive Gaussian noise of zero mean and a standard deviation of $\sigma_{noise}$, which is also defined as the power/energy of the noise. The value of $\sigma_{noise}$ of a noise added to a particular layer is selected using the signal to noise ratio ($SNR$) or noise form factor (${\eta}$) and the standard deviation of the weights in that layer as defined in equations (\ref{equ1}) - (\ref{equ3}). An additive Gaussian noise of zero mean and $\sigma_{noise}$ value equivalent to the noise form factor of 1\%, 5\%, 10\%, 20\%, 30\%, and 40\% is injected into all the model weights to evaluate the model noise-resistant property in this paper. It should be noted that the model baseline value, which is the model performance in the absence of noise, is equivalent to additive Gaussian noise of zero mean and ${\eta}$ value of zero, and it is denoted by $a_o$.
\begin{equation}
\label{equ1}
{\sigma_{noise}}=\frac{\sigma_w}{SNR}
\end{equation}
\begin{equation}
\label{equ2}
{\eta}=\frac{1}{SNR}
\end{equation}

\begin{equation}
\label{equ3}
{\sigma_{noise}}={\eta}\times{\sigma_w}
\end{equation}
\newcommand{\comment}[1]{}

\begin{figure*}[htbp]
\centering
\includegraphics[width=18cm]{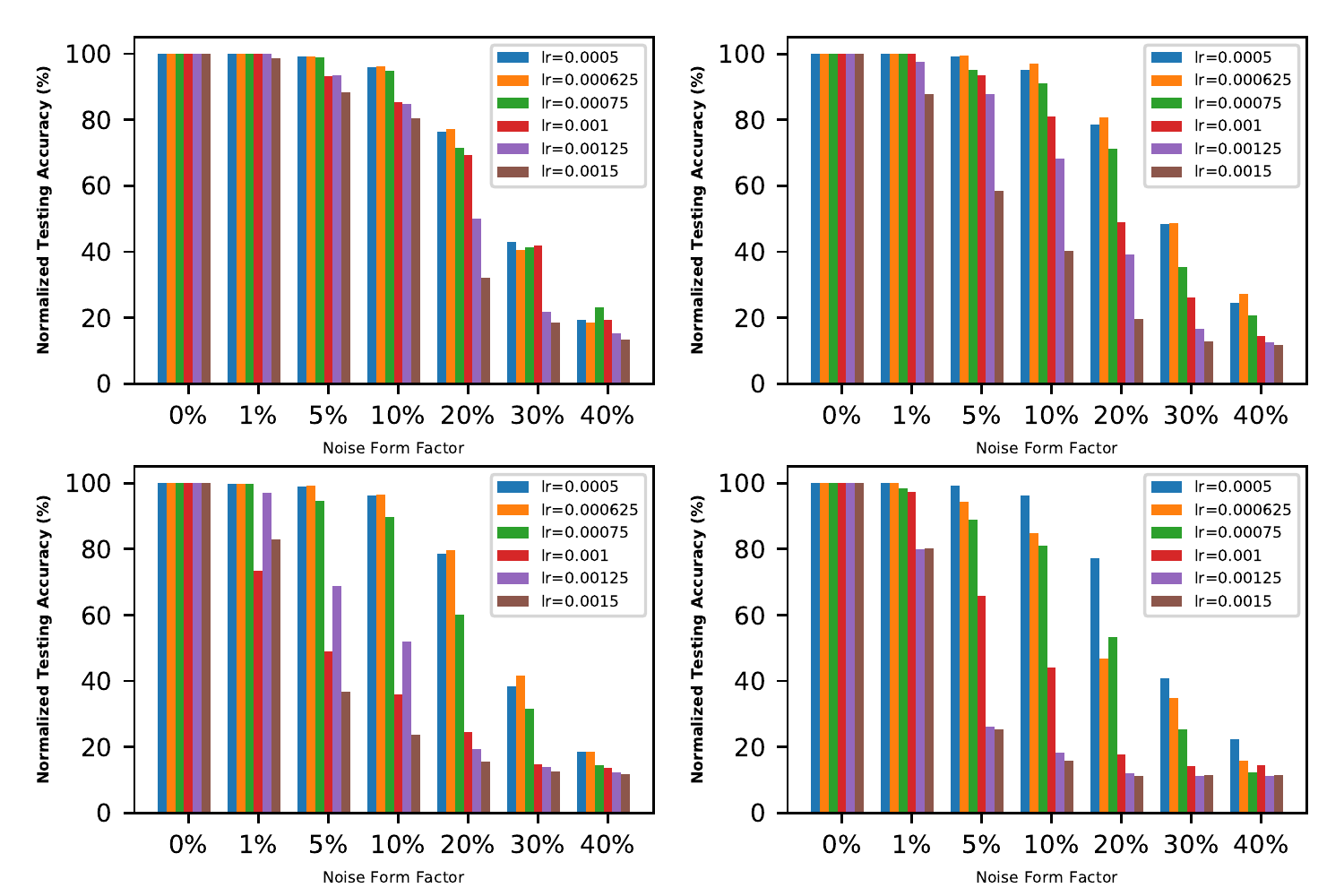}
\caption{The comparative study of noise resistant property of various models trained with CIFAR10 dataset with learning rate of values 0.0005, 0.000625, 0.00075, 0.001, 0.00125 and 0.00150, respectively. The noise resistant property is measured with Normalized Test Accuracy. The figure shows the plot of noise form factor against the Normalized Test Accuracy for ResNet20 (top left), ResNet32 (top right), ResNet44 (bottom left), and ResNet56 (bottom right), respectively. Noise form factor of 0 is equivalent to when no noise is injected into the models.}
\label{cifar10NormalizedImage}
\end{figure*}

After that, the impact of the noise is evaluated by measuring the new inference accuracy value of the DNN model on the test dataset. This is done for each of the six noise form factor values of 1\%, 5\%, 10\%, 20\%, 30\%, and 40\%, respectively, and for all the variants of each of the DNN models in Table~\ref{table:details}. The variants are because of the use of different learning rate values of 0.0005, 0.000625, 0.00075, 0.001, 0.00125, and 0.00150, respectively, to train each of the DNN models. It should be noted that the steps to calculate the inference accuracy due to the presence of white noise of $\sigma_{noise}$ value of $X$ is repeated multiple times due to the stochastic nature of the noise and average calculated as stated in equation (\ref{avg}) where $a_i$ is the inference accuracy at instant $i$ and $N$ is the number of times the experiment is performed. Furthermore, the inference accuracy due to the presence of noise is then normalized using the baseline inference accuracy value to get the normalized inference accuracy as stated in equation (\ref{equ4}) where ${A^x}$, ${a^x}$, and ${a}_o$ are the normalized classification accuracy, classification accuracy due to the presence of noise of $\eta$ value of $x\%$, and the baseline classification accuracy respectively. The normalized inference accuracy is essential as it is a fairer way of comparing the performance of the various DNN models due to noise as it captures the percentage change in model inference accuracy.

\begin{equation}
\label{avg}
{a^{x}}=\frac{\sum\limits_{i=1}^{N} a_i}{N}
\end{equation}

\begin{equation}
\label{equ4}
{A^x}=\frac{{a^x}} {{a}_o} \\
\end{equation}

\subsection{Software and Hardware}
\label{subsec:ExpSetup}
All the models used in this work are trained and tested using Keras deep learning framework installed on the NVIDIA V100-DGXS-32GB GPU.

\section{RESULTS and ANALYSIS}
\label{sec:results}

\begin{figure*}[htbp]
\centering
\includegraphics[width=18cm]{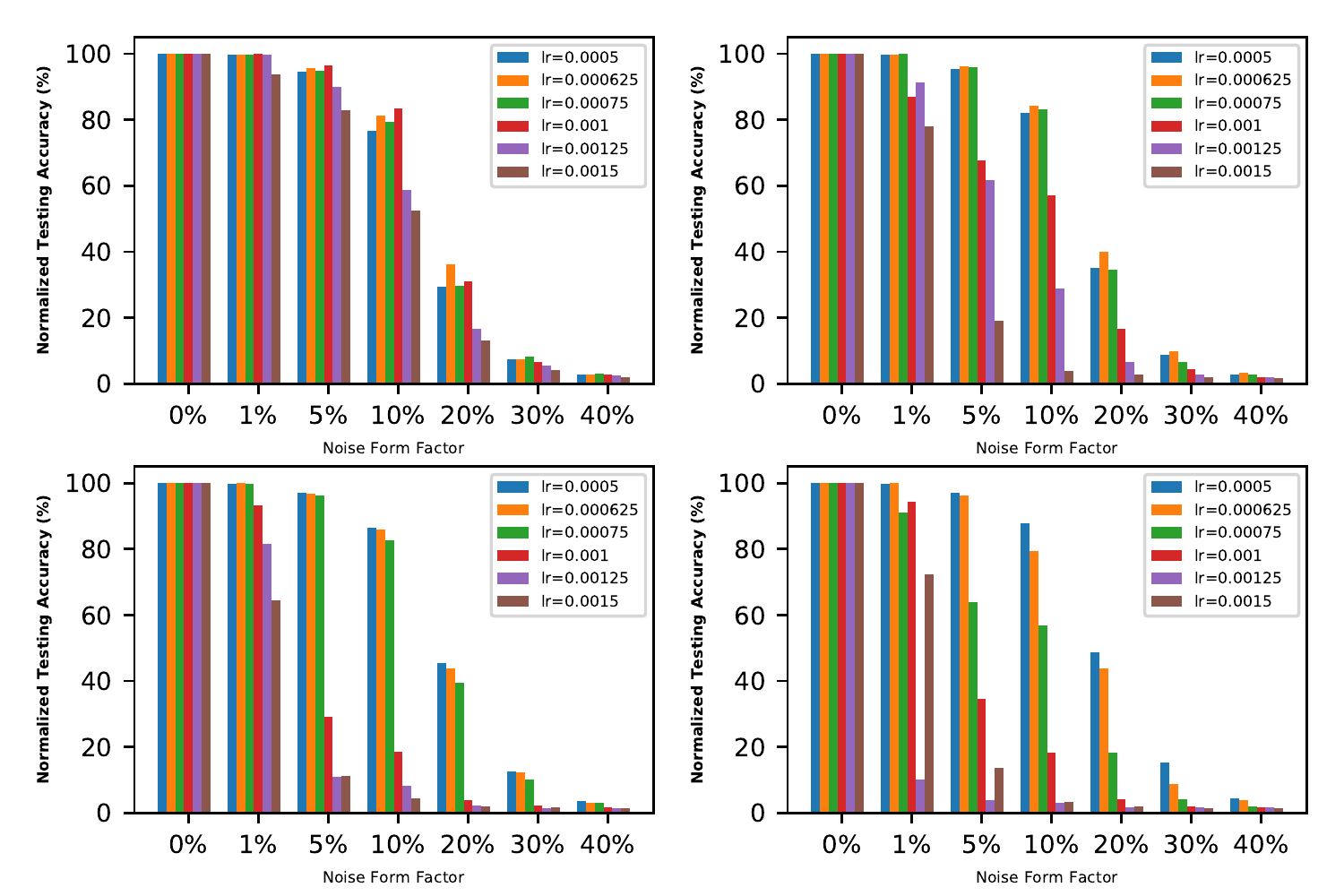}
\caption{The comparative study of noise resistant property of various models trained with CIFAR100 dataset with learning rate of values 0.0005, 0.000625, 0.00075, 0.001, 0.00125 and 0.00150, respectively. The noise resistant property is measured with Normalized Test Accuracy. The figure shows the plot of noise form factor against the Normalized Test Accuracy for ResNet20 (top left), ResNet32 (top right), ResNet44 (bottom left), and ResNet56 (bottom right), respectively. Noise form factor of 0 is equivalent to when no noise is injected into the models.}
\label{cifar100NormalizedImage}
\end{figure*}

The noise-resistant property of various models trained with different learning rate values in section \ref{sec:methods} is compared in this section. The noise-resistant property is evaluated by injecting analog noise into all the model layers and evaluating the resulting inference accuracy and the normalized inference accuracy.

\begin{table*}
\centering
\caption{Comparison of the average normalized inference accuracy of ResNet20, ResNet32, ResNet44, and ResNet56 models trained with learning rate values of 0.0005, 0.000625, 0.00075, 0.001, 0.00125 and 0.00150 with CIFAR10 and CIFAR100 dataset. The average normalized accuracy is defined in equation (\ref{equu}).}
\label{tab:result_summary}
\begin{tabular}{|c|c|c|c|c|c|c|c|c|c|c|c|c|c|c|c|c|c|c|c|c|c|}
\hline

{Dataset}& \multicolumn{6}{|c|}{CIFAR10}& \multicolumn{6}{|c|}{CIFAR100}\\ \hline

& \multicolumn{6}{|c|}{Learning Rate}& \multicolumn{6}{|c|}{Learning Rate}\\ \hline


{Model Name} & {0.0005} & {0.000625} & {0.000725}& {0.001} & {0.00125} & {0.0015} &
{0.0005} & {0.000625} & {0.000725}& {0.001} & {0.00125} & {0.0015}\\ \hline

{ResNet20} &{72.28\%} & {71.93\%} & {71.57\%}& {68.15\%} & {60.87\%} & {55.26\%} &
{51.77\%} & {53.81\%} & {52.50\%}& {53.30\%} & {45.50\%} & {41.36\%}\\ \hline

{ResNet32} & {74.36\%} & {75.53\%} & {68.90\%}& {60.67\%} & {53.72\%} & {38.48\%} &
{54.02\%} & {55.60\%} & {53.87\%}& {39.11\%} & {32.14\%} & {17.84\%}\\ \hline

{ResNet44} & {71.80\%} & {72.61\%} & {65.09\%}& {35.23\%} & {43.93\%} & {30.48\%} &
{57.52\%} & {56.97\%} & {55.20\%}& {24.79\%} & {17.67\%} & {14.17\%}\\ \hline

{ResNet56} & {72.67\%} & {62.81\%} & {59.81\%}& {42.24\%} & {26.47\%} & {25.93\%} &
{58.83\%} & {55.38\%} & {39.45\%}& {25.83\%} & {3.67\%} & {15.67\%}\\ \hline

\hline
\end{tabular}
\end{table*}

\subsubsection{Baseline Performance}
\label{sec:baseline}
Table \ref{tab:result_baseline} presents the baseline inference accuracy of the DNN models given in Table~\ref{table:details}. The baseline inference accuracy is obtained when evaluating trained DNN models on the testing dataset without noise, equivalent to injecting with Gaussian noise of zero mean and zero standard deviation. The baseline inference accuracy of a DNN model trained on CIFAR10 is higher than that of the corresponding DNN model trained on CIFAR100. The higher number of classes and the classes' overlapping nature mean that the classification task on the CIFAR100 dataset is more complex than the one on the CIFAR10 dataset. Furthermore, the reduction in the number of data per class for the CIFAR100 dataset and the small dimensions of the image means that the performance of the DNN models is limited.

Also, It can be observed that for both CIFAR10 and CIFAR100 datasets, the use of lower or higher value of the learning rate did not substantially give any performance advantage to the trained DNN models. For example, the ResNet20 model has a baseline inference value of 91.11\%, 91.39\%, 91.87\%, 91.27\%, 91.53\% and 91.49\% for learning rate value of 0.0005, 0.000625, 0.00075, 0.001, 0.00125 and 0.00150 respectively. A similar observation is also made for ResNet20 trained on the CIFAR100 dataset with a baseline inference value of 65.22\%, 66.05\%, 66.79\%, 66.89\%, 67.37\%, and 68. 37\% for learning rate values of 0.0005, 0.000625, 0.00075, 0.001, 0.00125 and 0.00150 respectively. It must be noted that a higher learning rate confers an advantage in that the training process is faster, although the weight might end up at the saddle point or the local minimum. This observation is also true for ResNet33, ResNet44 and ResNet56 models.

\subsubsection{Performance in the presence of Analog noise}
\label{sec:noise_performance}

The impact of the learning rate on the noise-resistant property of a DNN model is analyzed by comparing the normalized inference accuracy of the six variants of the same DNN model architecture obtained when trained with different learning rates. The normalized learning rate is used to fairly capture the percentage change in model inference accuracy due to the noise. Figures \ref{cifar10NormalizedImage} and \ref{cifar100NormalizedImage} show the normalized inference accuracy due to the injected noise for ResNet20, ResNet32, ResNet34, and ResNet56 models trained with different learning rate as stated in section \ref{sec:training} using CIFAR10 and CIFAR100 datasets, respectively. It can be observed that the normalized inference accuracy, which is a measure of the noise-resistant property of the DNN model, decreases with the noise form factor, which is a measure of the power of the injected noise. Furthermore, the degradation in the model's performance also occurred gracefully, speaking to the noise-resistant property of DNN models. This observation is the same irrespective of the model's type, the classification task's complexity, and the learning rate used in training the model, although the degradation rate varies. The observation can be attributed to the stronger tendency of the noise to corrupt the model weight due to its higher magnitude, which increases with an increase in noise power/noise form factor. For a variant of ResNet56 trained with a learning rate of 0.0.00075 on the CIFAR100 dataset, normalized accuracy value of 98.34\%, 88.81\%, 80.91\%, 53.32\%, 25.25\%, and 12.27\% is obtained when the model is injected with the noise of noise form factor of 1\%, 5\%, 10\%, 20\%, 30\%, and 40\%, respectively. A normalized accuracy value of 91.12\%, 63.87\%, 56.98\%, 18.33\%, 4.29\%, and 2.09\% is also obtained when the same model, trained and tested with CIFAR100 dataset, is injected with the noise of noise form factor of 1\%, 5\%, 10\%, 20\%, 30\%, and 40\%, respectively.

\begin{figure}[htbp]
\centering
\includegraphics[width=9.5cm]{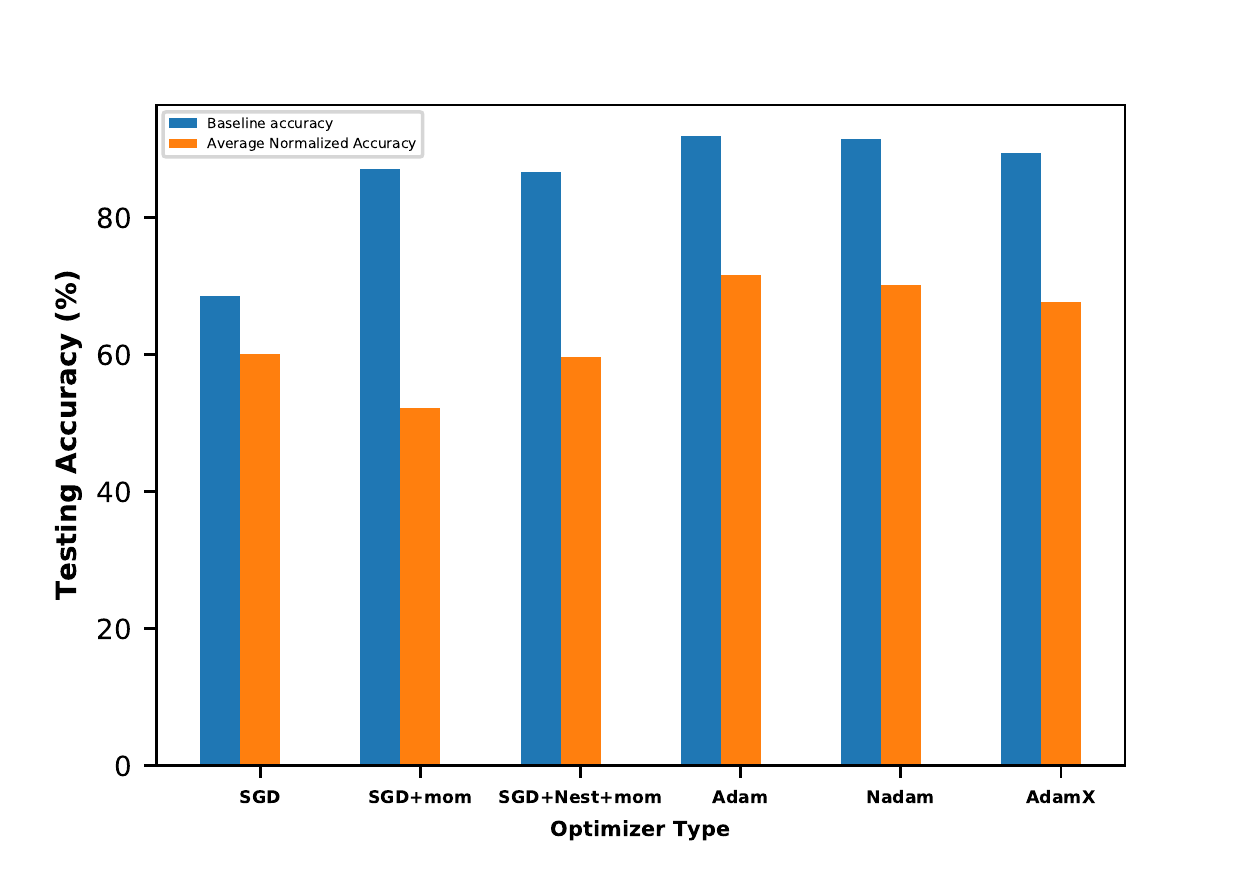}
\caption{ Comparative study of the effect of different optimizers (with or without adaptive learning rate) on model inference performance and model noise-resistant property using ResNet20 trained on CIFAR10 dataset. The optimizers used are SGD, SGD with momentum, SGD with Nesterov momentum, and Adam, Nadam, and AdamX algorithms. The model inference performance is measured using the baseline inference accuracy, and the model noise-resistant property is measured using the average normalized accuracy as defined in equation (\ref{equu}). }
\label{adaptive}
\end{figure}

\subsubsection{Impact of Different Learning Rates with Adam Optimizer}
\label{sec:BatchNorm}
The noise-resistant property of the variants of each of the DNN models, under consideration in this work, is compared in this section. The performance of the variants, obtained by training the same model architecture with different learning rates, when injected with additive Gaussian noise and measured using the normalized inference accuracy is shown in Figures \ref{cifar10NormalizedImage} and \ref{cifar100NormalizedImage}. It can be observed that the noise-resistant property of a DNN model of specific architecture is affected by the learning rate. The observation holds for all the DNN models irrespective of learning rate values and datasets the models are trained on. For example, a normalized inference accuracy value of 95.93\%, 96.17\%, 94.78\%, 85.35\%, 84.93\%, and 80.52\% is obtained for ResNet20 models trained on CIFAR10 dataset with learning rate of 0.0005, 0.000625, 0.00075, 0.001, 0.00125 and 0.00150 respectively when injected with Gaussian noise of noise form factor of 10\%. This trend is also observed for variants of ResNet20 DNN model trained on CIFAR100 dataset trained with learning rates of 0.0005, 0.000625, 0.00075, 0.001, 0.00125, and 0.00150, which has normalized inference accuracy values of 76.55\%, 81.26\%, 79.43\%, 83.33\%, 58.78\%, and 52.48\% when injected with additive Gaussian noise of noise form factor of 10\%. This shows that the noise-resistant property of a DNN model can be tuned by varying the learning rate without compromising too much on the model performance.

A new metric, average normalized inference accuracy, defined in equation (\ref{equu}) is used to summarize the performances of the variant of each model in the presence of noise and stated in Table \ref{tab:result_summary} where $A_i$ is the normalized inference accuracy at a noise form factor, and $N$ is the number of non-baseline noise factor which is 6.

\begin{equation}
\label{equu}
A_{avr}=\frac{\sum\limits_{i=1}^{N} A_i}{N}
\end{equation}

Table \ref{tab:result_summary} shows that the learning rate that gives the best performance in the presence of analog noise varies from one model to another. While the learning rate value of 0.0005 gives the best noise-resistant property for ResNet20 and ResNet56 trained on the CIFAR10 dataset, the learning rate value of 0.000625 gives the best noise-resistant property for ResNet32 and ResNet44 trained on the CIFAR10 dataset. This observation also applies to these models trained on the CIFAR100 dataset, where the learning rate value of 0.000625 gives the best performance for ResNet20 and ResNet32, and the learning rate of 0.0005 gives the best noise-resistant property for ResNet44 and ResNet56.

\begin{figure*}[htbp]
\centering
\includegraphics[width=18cm]{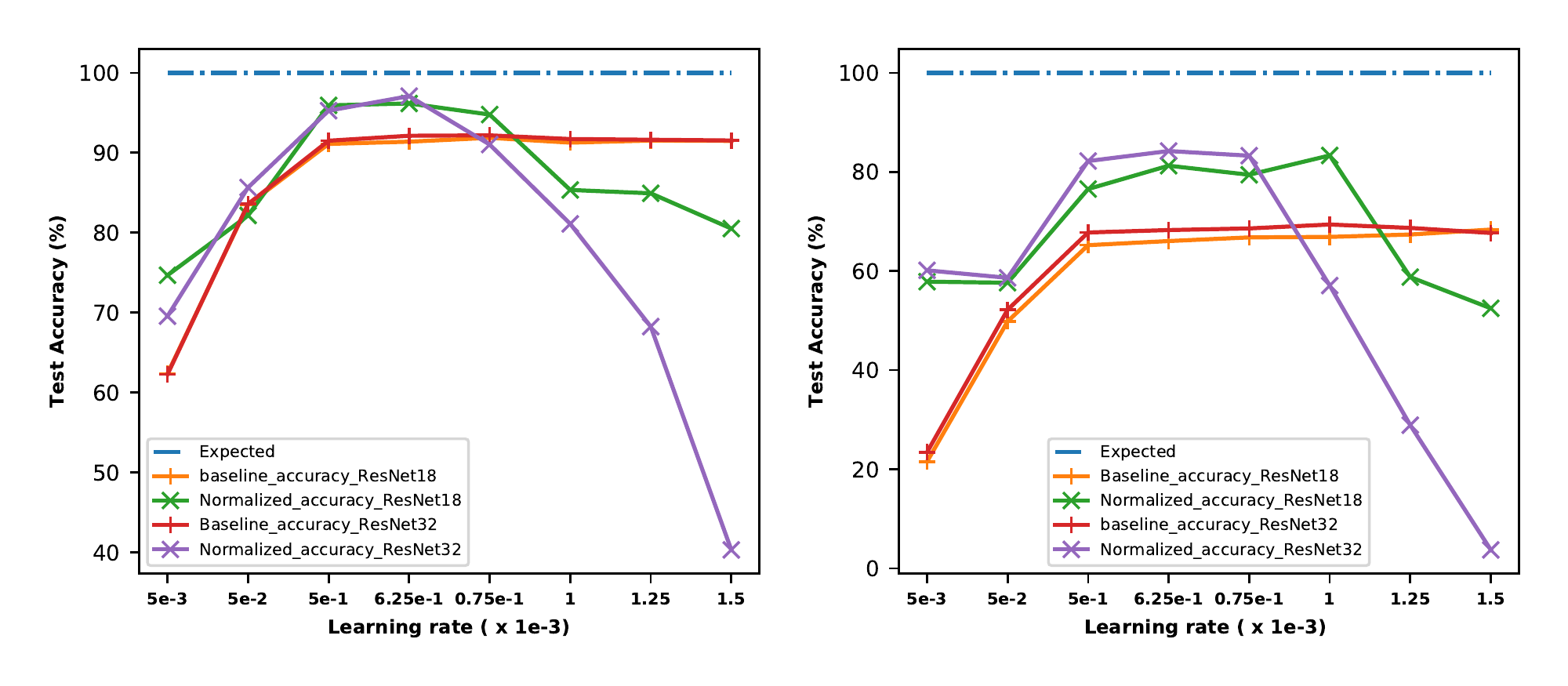}
\caption{The plot shows the tradeoff between model inference performance and its noise-resistant property as the learning rate is increased for CIFAR10 (left) and CIFAR100 (right) for ResNet20 and ResNet32. The model inference accuracy is measured using the baseline inference accuracy, and the model noise-resistant property is measured using the average normalized accuracy as defined in equation (\ref{equu}).}
\label{tradeoff}
\end{figure*}

\subsubsection{Impact of Different Optimizers (with or without adaptive learning rate)}
\label{sec:BatchNorm}
A comparison study of the noise-resistant property of DNN models trained with adaptive learning rate and fixed global learning rate is done in this section. This is done by training the ResNet20 model with a global learning rate of 0.0.00075 using SGD, SGD with momentum, SGD with Nesterov momentum, Adam, Nadam, and AdamX optimizer. The Adam, Nadam, and AdamX optimizers are types of adaptive learning rate, as they use localized learning rate for each parameter during training compared with SGD, SGD with momentum, and SGD with Nesterov momentum that uses fixed and equal learning rates for all parameters. The result of our comparative study is shown in Figure \ref{adaptive} which shows a plot of baseline inference accuracy and average normalized accuracy for ResNet20 model trained using a global learning rate value of 0.0.00075 using the selected optimizers. It can be observed that the different optimizers do affect both the model inference performance and its noise-resistant property. The result showed that the ResNet20 model has a baseline inference accuracy value of 68.51\%, 87.09\%, 86.65\%, 91.87\%, 91.57\%, 89.47\% using SGD, SGD with momentum, SGD with Nesterov momentum, Adam, Nadam and AdamX optimizers, respectively. Average normalized accuracy of 52\%, 60\%, 60\%, 72\%, 70\%, and 68\% are also obtained for the resulting ResNet20 model trained using SGD, SGD with momentum, SGD with Nesterov momentum, Adam, Nadam, and AdamX optimizers, respectively. The result shows that the models with adaptive learning rates have better model inference performance and noise-resistant property than models with fixed learning. The improved performance might be due to the dependence of the effective local learning rate in the adaptive learning rate algorithm on the past learning rate, which increases the learning rate used for training. This means that the power of the gradient noise present during training is higher for adaptive learning rate scenarios as the resulting learning rate influences the power of the noise.

\subsubsection{Tradeoff between Prediction Accuracy and Model Noise Resistant Property}
\label{sec:BatchNorm}
The plot of baseline inference accuracy for ResNet20 in the absence of noise and normalized inference accuracy for ResNet20 due to the injection of the noise of noise form factor of 10\% for learning rate values of 0.0005, 0.000625, 0.00075, 0.001, 0.00125, and 0.00150 is shown in Figure~\ref{tradeoff}. The table shows that at a learning rate value of 5e-5 and lower, the model inference performance measured using the baseline inference accuracy and the noise-resistant property when injected with the noise of noise factor of 10\% is adversely affected. This learning rate region is not desirable even if the noise-resistant property is perfect as the baseline inference performance is poor. At a learning rate value of 1e-3 and above, improved inference performance is obtained as measured by the baseline inference accuracy. However, the noise-resistant property of the model as measured by the normalized inference accuracy trends downward with an increase in the learning rate. This region is desirable if an excellent noise-resistant property is not desired as the baseline inference accuracy value is good. However, the need for excellent noise-resistant property for analog devices means that this region is not desirable as this property is the most desirable as training models in this region might be faster. The region with great noise-resistant property and excellent inference performance are the most desirable for analog hardware as this region finds a balance between performance and noise-resistant property. This desired region is the region with a learning rate value between 0.0005 and 0.001 in Figure \ref{tradeoff} for ResNet20 as the baseline inference accuracy, and normalized inference accuracy is high in this region, finding a balance between performance and noise-resistant property is achieved. The three regions establish that changing the learning rate also impacts the noise-resistant property and inference performance of the DNN model. Hence, there is a need to select the learning rate value in the region where there is a balance between model performance and model noise-resistant property.

\subsubsection{Impact of Task Complexity on Model Noise Resistant Property}
\label{sec:complexity}

Table \ref{tab:result_summary} showed that the average normalized inference accuracy of DNN models due to the injection of noise into the model weight is affected by the complexity of the task. The table showed that ResNet20, ResNet32, ResNet44, and ResNet56 trained on the CIFAR10 dataset with a learning rate of 0.000625 has an average normalized inference accuracy of 71.93\%, 75.53\%, 72.61\%, and 62.81\%, respectively. These values are higher than 53.81\%, 55.06\%, 56.97\%, and 55.38\% normalized inference accuracy values obtained for the same models trained on the CIFAR100 dataset. This performance gap is noticed irrespective of the DNN model type and the learning rates the models are trained with. These show that the complexity of the model's task strongly influences its performance in the presence of Gaussian noise as the CIFAR100 dataset is a more complex classification task than CIFAR10. The complexity of the CIFAR100 dataset is because it contains 100 classes, with most of the classes overlapping compared to the ten non-overlapping classes in the CIFAR10 dataset.

\section{Related Works and Discussions}
\label{sec:discussion}
Learning rate is the most critical hyperparameter for stochastic gradient-based optimization problems as it influences if convergence is possible and the rate of convergence. It works by influencing the size of the steps to take in the direction of the negative gradient of the parameters to be estimated. Therefore, there is a need to find an effective method to select and design an appropriate value for the learning rate at each iteration. This need has given rise to many heuristics for estimating a reasonable learning rate at each iteration by either increasing the learning rate when suitable or decreasing the learning rate when near a local minimal~\cite{zeiler2012adadelta}. For example, the authors in \cite{Robbins2007ASA} introduced a learning rate annealing method, a method that automatically reduces the size of the learning rate based on the number of epoch. It is also possible to decrease the learning rate when the validation accuracy appears to have plateaus~\cite{zeiler2012adadelta}. These approaches are critical to prevent the parameter values from oscillating back and forth around the minimal. Also, there is a recent paradigm that involves allowing the global learning rate to cyclically vary between reasonable boundary values during training~\cite{smith2017cyclical}. This method automatically eliminates the need to tune the learning rate and needs no extra calculation. In recent times, there has been an introduction of adaptive learning rate~\cite{schaul2013pesky,gulcehre2017robust} such as Adam~\cite{kingma2017adam}, Adadelta~\cite{zeiler2012adadelta}, RMSprop, etc. instead of a global learning rate, and this new method can be used in combination with the Cyclical Learning Rates. The need for an adaptive learning rate is because different dimensions in the model parameter vectors interact with the cost function in entirely different ways, and using a localized learning rate for this dimension can significantly improve the training process. Also, the authors in ~\cite{smith2018superconvergence} demonstrated the use of a large learning rate to achieve improved speed and performance. The authors argued that a large learning rate provides some form of regularization for model training when other forms of regularization are reduced to maintain an optimal balance in regularization. The work in~\cite{jastrzebski2018factors} also showed that learning rate in conjunction with the batch size and gradient covariance influences the minima found by SGD during model training.

The need to design models with excellent noise-resistant properties for deployment on analog accelerators has led to studies on noise-resistant properties of popular DNN models and also novel methods to design and train models with excellent noise-resistant properties~\cite{Fagbohungbe2020BenchmarkingIP,joshi}. The authors in ~\cite{joshi} and ~\cite{Kariyappa} proposed the use of additive and multiplication noise injection methods, respectively, to increase the noise-resistant property of DNN models. These noise injection methods are also combined with knowledge distillation to achieve even better noise-resistant property in~\cite{NoisyNN}. Also, the impact of Batch Normalization on the noise-resistant property of DNN models is investigated, and a novel DNN design and training method that finds a balance between model inference performance and model noise-resistant property is proposed in \cite{BatchNorm2020}. A comparative performance of various BatchNorm types on the model noise-resistant property is also done in \cite{bay1}. A method that conditions DNN models by exposing them to a noisy computation environment are proposed in~\cite{Bo,schmid} and a chip in the loop method that adapts pre-trained model weights for the inference only chip is introduced in~\cite{NeuroSmith}. The use of generalized fault aware pruning technique to improve model resilience to hardware fault is also discussed in~\cite{Zhang2019}. The use of error correction code~\cite{Upadhyaya2019ErrorCF} only or in combination with other methods such as reinforcement learning~\cite{huang2020functional} can also be used to ensure models are resistant to fault during computation.

This work is different from existing work on learning rate as it does not propose new ways of selecting learning rate values~\cite{schaul2013pesky} or design learning rate for model training~\cite{smith2017cyclica,kingma2017adam}. To the best of our knowledge, it is the first paper to investigate the impact of the learning rate on the noise-resistant property of DNN models. This work is significant as the noise-resistant property of DNN models can be tuned by increasing/decreasing the learning rate. This paper shares some similarities with~\cite{joshi, NoisyNN} in that computational noise is modeled as analog noise. It is still very different as this research did not explore the use of external analog noise to improve the DNN model noise-resistant property as done in those works. Furthermore,  the insight from this work can help select appropriate learning rate values that maintain the delicate balance between model inference performance and its noise-resistant property. It is also very different from the work in ~\cite{Upadhyaya2019ErrorCF,Zhang2019,huang2020functional} where model noise is modeled as digital noise.

\section{{Conclusion}}
\label{sec:conclusion}
The impact of the learning rate on the analog noise-resistant property of DNN models is studied in this work. The finding is significant because the learning rate is a critical model training hyperparameter. The study is achieved by training DNN models with different learning rates and comparing the resulting models' inference performances when analog noise is injected (additive white Gaussian noise) into the model parameters.
The experimental results demonstrate that the choice of learning rate value and the optimizer type significantly impact the noise-resistant property of DNN models. Specifically, this work established a range of learning rate values that maintain the delicate balance between model prediction accuracy and model noise-resistant property. At a learning rate value below this range of values, the model inference performance and noise resistant property are adversely affected. At a learning rate value above this range, the resulting model has excellent model inference performance but poor noise-resistant property. In addition, through theoretical analysis of the model training process, it is clear that the impact of learning rate on the model noise-resistant property is due to its influence on the inherent noise present in the model training process, which is responsible for the noise-resistant property of DNN models.

\section{Acknowledgments}
\label{acknowledgement}
This research work is supported by the U.S. Office of the Under Secretary of Defense for Research and Engineering (OUSD(R\&E)) under agreement number FA8750-15-2-0119. The U.S. Government is authorized to reproduce and distribute reprints for governmental purposes, notwithstanding any copyright notation thereon. The views and conclusions contained herein are those of the authors and should not be interpreted as necessarily representing the official policies or endorsements, either expressed or implied, of the Office of the Under Secretary of Defense for Research and Engineering (OUSD(R\&E)) or the U.S. Government.

\bibliographystyle{IEEEtran}

\bibliography{BatchNormalisation}

$ $ \\

\end{document}